\documentclass[times, review, 10pt]{elsarticle}
\usepackage{amsmath,amsfonts}
\usepackage{algorithmic}
\usepackage{array}
\usepackage{textcomp}
\usepackage{stfloats}
\usepackage{url}
\usepackage{verbatim}
\usepackage{graphicx}
\usepackage{subcaption}
\usepackage{bm}
\usepackage{float} 
\usepackage{booktabs}
\usepackage{array, caption, threeparttable}
\usepackage[font=small,labelfont=bf,labelsep=none]{caption}
\usepackage{balance}
\usepackage{algorithm}
\usepackage{natbib}
\setcitestyle{numbers,square}
\usepackage{color}
\usepackage{stfloats}
\newtheorem{Definition}{\textbf{Definition}}
\makeatletter

\usepackage{hyperref}
\journal{}
\makeatother

\begin{document}
	
\begin{frontmatter}
	
\title{Binarization-Aware Adjuster for Discrete Decision Learning with an Application to Edge Detection}

\author{Hao Shu\corref{cor1}}
\ead{Hao_B_Shu@163.com}
\cortext[cor1]{}

\affiliation{organization={Sun-Yat-Sen University},
	city={Shenzhen},
	country={China}}
\affiliation{organization={Shenzhen University},
	city={Shenzhen},
	country={China}}

\begin{abstract}

Discrete decision tasks in machine learning exhibit a fundamental misalignment between training and inference: models are optimized with continuous-valued outputs but evaluated using discrete predictions. This misalignment arises from the discontinuity of discretization operations, which prevents decision behavior from being directly incorporated into gradient-based optimization. To address this issue, we propose a theoretically grounded framework termed the Binarization-Aware Adjuster (BAA), which embeds binarization characteristics into continuous optimization. The framework is built upon the Distance Weight Function (DWF), which modulates loss contributions according to prediction correctness and proximity to the decision threshold, thereby aligning optimization emphasis with decision-critical regions while remaining compatible with standard learning pipelines. We apply the proposed BAA framework to the edge detection (ED) task, a representative binary decision problem. Experimental results on representative models and datasets show that incorporating BAA into optimization leads to consistent performance improvements, supporting its effectiveness. Overall, this work establishes a principled approach for aligning continuous optimization with discrete decision behavior, with its effectiveness demonstrated in a concrete application setting.

\end{abstract}




\end{frontmatter}

\section{Introduction}
\label{Introduction}

Many real-world vision tasks, such as semantic segmentation, image classification, object recognition, and edge detection (ED)~\cite{NN2019EdgeConnect,MR2011Edge,ZD2007An,K1983On,C1986A}, ultimately involve discrete decision processes (e.g., binary or categorical outputs). However, modern learning methods, including deep neural networks, are typically optimized using continuous-valued loss functions within gradient-based learning frameworks due to the requirement of differentiable computation~\cite{PH2021RINDNet,PH2022Edter,ZH2024Generative,YX2024Diffusionedge,S2025Boost}. As non-differentiable inference operations, such as thresholding, quantization, or category assignment, are excluded from the optimization process, a practical inconsistency arises between continuous-valued optimization during training and discrete inference at test time. This inconsistency may lead to suboptimal learning behavior, as the optimization objective does not explicitly reflect decision-critical outcomes.

This issue is particularly evident in pixel-level binary prediction tasks. For example, edge detection (ED) aims to generate binary edge maps but is typically trained using continuous-valued loss functions such as binary cross-entropy (BCE). Since binarization is non-differentiable, it is excluded from the optimization loop and applied only during inference and evaluation. Consequently, model updates tend to emphasize predictions that are easy to optimize but have limited impact on the final binary output, while predictions near the decision boundary, which are more critical for classification accuracy, often receive insufficient attention due to their lower contribution to the loss and higher optimization difficulty.

Several studies have attempted to alleviate the discrepancy between training objectives and decision behavior, typically by modifying loss functions to rebalance decision classes, such as weighted BCE (WBCE)~\cite{XT2015Holistically}, or by adopting alternative loss formulations~\cite{CK2024RankED,S2025Rethinking}. While these methods improve model performance, they do not explicitly model the influence of binarization on the optimization process. As a result, the underlying misalignment between continuous optimization and discrete inference remains largely unaddressed.

To address this issue, we propose a theoretically motivated framework termed the \textit{Binarization-Aware Adjuster (BAA)}, which incorporates binarization-related decision characteristics into gradient-based optimization in a principled and continuous manner. The core idea of the BAA is a threshold-aware loss adjustment mechanism based on the proposed \textit{Distance Weight Function (DWF)}, which reweights pixel-level loss contributions according to prediction correctness and proximity to the decision threshold. This formulation is designed to emphasize decision-critical predictions near the binarization boundary while reducing the influence of confidently correct predictions whose further optimization has limited effect on the final discrete outcome. Since the BAA is continuous, it can be directly integrated into standard training pipelines without altering model architectures or inference procedures.

The proposed framework focuses on binary decision learning and is evaluated on the ED task as a representative application. ED is chosen due to its fundamental role in low-level vision and its strong dependence on precise binary decisions at the pixel level, making it well-suited for analyzing the interaction between continuous optimization and discrete inference. While the present study concentrates on ED, the formulation of the BAA is general in nature and provides theoretical insight into how decision boundaries can be more effectively reflected in continuous optimization objectives.

The main contributions of this work are summarized as follows:

\begin{itemize}
\item We analyze a practical inconsistency between continuous-valued optimization and discrete inference in gradient-based learning and formalize its impact on binary prediction tasks.
\item We propose the Binarization-Aware Adjuster (BAA), a principled and continuous loss adjustment framework based on the Distance Weight Function (DWF), to incorporate binarization behavior into the optimization process.
\item We validate the proposed framework on the edge detection (ED) task through extensive experiments on multiple benchmark datasets and architectures.
\end{itemize}

\section{Previous Works}

A large body of learning-based systems perform discrete decision-making at inference while being optimized using continuous-valued loss functions during training. This discrepancy has motivated extensive research on improving the alignment between continuous optimization objectives and discrete evaluation criteria. In this work, edge detection (ED) is adopted as a representative validation scenario due to its strong dependence on binary decisions and its close connection to a wide range of structured prediction problems. In the following, we review relevant studies using ED as an illustrative example, with a focus on datasets, model architectures, loss functions, and evaluation protocols.

\subsection{Datasets}

Early research in computer vision commonly relied on general-purpose datasets. For instance, BSDS300/500~\cite{MF2001A} have been widely used in contour detection, segmentation, and ED. Similarly, datasets such as PASCAL VOC~\cite{EV2010The}, COCO~\cite{LM2014Microsoft}, and NYUD~\cite{SH2012Indoor} have supported a variety of tasks including segmentation, detection, and image classification. As research topics became more specialized, dataset construction increasingly focused on task-specific characteristics, e.g., PASCAL VOC for object detection, NYUD for indoor segmentation, and BSDS500 for contour-based analysis.

For ED in particular, several dedicated datasets have been developed to provide more accurate and task-oriented annotations. BIPED and BIPED2~\cite{SR2020Dense} provide high-resolution edge annotations for campus scenes, BRIND~\cite{PH2021RINDNet} extends BSDS500 with refined edge labels, and UDED~\cite{SL2023Tiny} offers carefully curated samples with improved image and annotation quality.

Despite these efforts, annotation quality remains a persistent challenge. Human-labeled edge maps often exhibit ambiguity, inconsistency, and noise, especially near weak or perceptually subjective boundaries. Although recent studies have proposed robust training strategies and improved annotation protocols~\cite{FG2023Practical, WD2024One, S2025Enhancing}, annotation noise continues to affect both model generalization and the reliability of quantitative evaluation.

\subsection{Model Architectures}

Deep learning architectures have achieved remarkable success across a wide range of domains, including computer vision~\cite{RR2020ESRGAN}, natural language processing~\cite{VS2017Attention}, and decision-making systems~\cite{CZ2024Versatile}. Representative architectural paradigms, such as convolutional neural networks (CNNs) and transformers, have demonstrated strong modeling capacity and broad applicability.

In the context of ED, the seminal work HED~\cite{XT2015Holistically} demonstrated the effectiveness of multi-scale deep supervision within CNN architectures, significantly outperforming traditional edge detectors~\cite{K1983On,C1986A} and earlier feature-based learning approaches~\cite{MF2004Learning,LZ2013Sketch,AM2011Contour,DZ2015Fast,R2008Multi}. Subsequent CNN-based methods, including RCF~\cite{LC2017Richer}, BDCN~\cite{HZ2022BDCN}, PiDiNet~\cite{SL2021Pixel}, and DexiNet~\cite{SS2023Dense}, further refined network structures and feature fusion strategies. Transformer-based~\cite{PH2022Edter,JG2024EdgeNAT} and diffusion-based~\cite{ZH2024Generative,YX2024Diffusionedge} approaches have also been explored to capture long-range dependencies and generative priors. More recently, the Extractor–Selector framework~\cite{S2025Boost} introduced a modular design to improve feature selection. Despite architectural diversity, CNN-based models remain the most widely adopted in ED due to their robustness, efficiency, and mature training paradigms.

\subsection{Loss Functions}

Loss function design plays a central role in model optimization, particularly for binary decision tasks. BCE was initially the standard choice; however, it assigns equal importance to positive and negative samples, which is suboptimal for the highly imbalanced data distributions commonly encountered, such as in ED. To address this issue, WBCE has become a widely adopted baseline and remains one of the most standard loss formulations in practice.

Beyond WBCE, various alternative loss functions have been proposed to better reflect structural or ranking-based objectives. Examples include Dice loss~\cite{DS2018Learning} and tracing loss~\cite{HX2022Unmixing} for geometry-aware optimization, AP-loss~\cite{CL2019Towards} and RankED~\cite{CK2024RankED} based on ranking formulations, and SWBCE~\cite{S2025Rethinking}, which explicitly models uncertainty in human annotations. While these approaches improve class balance, order cues, or perceptual quality, they generally treat binarization as a post-optimization operation and do not explicitly account for the inference rule during gradient-based learning. Consequently, the continuous–discrete inconsistency persists, motivating the development of the proposed BAA framework.

\subsection{Evaluation Metrics}

For binary decision tasks, performance is typically evaluated using the F$\beta$ score:
\begin{equation}
F_\beta = \frac{(1+\beta^2) \times \text{Precision} \times \text{Recall}}{\beta^2 \times \text{Precision} + \text{Recall}},
\end{equation}
where
\begin{equation}
\text{Precision} = \frac{TP}{TP+FP}, \quad
\text{Recall} = \frac{TP}{TP+FN}.
\end{equation}
Here, $TP$ denotes true positives, $FP$ false positives, and $FN$ false negatives, with $\beta$ typically set to 1. The most commonly reported metrics include:

\begin{itemize}
\item \textbf{ODS (Optimal Dataset Scale):} The best F$_1$ score achieved using a single optimal threshold across the dataset.
\item \textbf{OIS (Optimal Image Scale):} The average of image-wise best F$_1$ scores obtained using image-specific optimal thresholds.
\end{itemize}

For ED, the evaluation protocol with spatial tolerance was originally introduced in~\cite{MF2004Learning} to account for localization deviations. More recent studies~\cite{S2025Enhancing} advocate stricter and standardized tolerance settings to improve the reliability and comparability of quantitative results.

\section{Methodology: General Theory}
\label{Methodology}

This section presents the theoretical framework of the proposed method. For clarity, the formulation is developed for binary decision tasks, which serve as a representative and intuitive setting. The discussion naturally extends to multi-valued discrete decision problems without altering the core principles.

\subsection{Motivation and Insight}

A fundamental challenge in training binary decision models lies in the inconsistency between training objectives and evaluation protocols. During training, model outputs are continuous-valued, typically normalized to $[0,1]$, and optimized using pixel-wise loss functions such as WBCE. In contrast, evaluation metrics are computed on binarized outputs obtained by applying a fixed threshold. Since binarization is a discontinuous operation, it cannot be directly incorporated into gradient-based optimization. As a result, the loss function does not explicitly reflect the discrete decisions that ultimately determine evaluation performance.

This inconsistency leads to an inefficiency in optimization. Predictions that are already on the correct side of the decision threshold and far from it, namely, confidently correct elements, tend to dominate the gradient signal because they are easy to optimize and contribute substantially to the loss. In contrast, predictions near the decision boundary, which are more ambiguous and critical for correct classification, are harder to optimize and therefore contribute less to the gradients. Consequently, the optimizer is implicitly biased toward improving easy but decision-irrelevant predictions, while allocating insufficient learning capacity to ambiguous yet decision-critical ones. This imbalance can prevent the model from converging to parameters that are optimal with respect to the final binary decisions.

As an illustrative example, Table~\ref{Thr} reports optimal binarization thresholds for several ED models trained with WBCE across multiple datasets. These thresholds typically lie between $0.6$ and $0.8$. Suppose the threshold is set to $0.7$. Reducing the prediction of a non-edge pixel from $0.50$ to $0.01$ can significantly decrease the loss, yet it has no effect on the final binary edge map. Such predictions are often optimized early in training and tend to dominate the optimization process due to their large quantity and easily reducible loss values. In contrast, increasing the prediction of an edge pixel from $0.69$ to $0.71$ directly alters the binary decision outcome, but is generally more difficult to optimize as a pixel with the predicted value close to the decision threshold. Moreover, these pixels typically appear near true edge regions, which are more critical for accurate discrimination, but are relatively sparse in number. As a result, their overall contribution to the loss function is limited, making such decision-critical predictions more likely to be under-emphasized during training.

\begin{table}[htbp]
\renewcommand\arraystretch{1.3}
\centering
\caption{\textbf{Optimal binarization thresholds under WBCE training for ED models}: Evaluation uses 1-pixel error tolerance without NMS. Experimental settings follow the ones in Section \ref{Settings}. Models are retrained on the pre-training set and evaluated on the validation set.}
\label{Thr}
\begin{tabular}{|p{20mm}<{\centering}|p{15mm}<{\centering}|p{15mm}<{\centering}|p{15mm}<{\centering}|p{15mm}<{\centering}|}
\hline
   &  UDED & BSDS500 & BRIND & BIPED2\\
\hline
HED & 0.74 & 0.85 & 0.77 & 0.79 \\
\hline
BDCN & 0.65& 0.79 & 0.78 & 0.72 \\
\hline
Dexi & 0.63 & 0.87 & 0.79 & 0.79 \\
\hline
EES3-Double & 0.61 & 0.86 & 0.77 & 0.80 \\
\hline
\end{tabular}
\end{table}

To address this issue, we propose a loss adjustment mechanism realized by the \textit{Binarization-Aware Adjuster (BAA)}, which dynamically reweights pixel-wise loss contributions based on both classification correctness and proximity to the decision threshold. By explicitly incorporating binarization-related decision characteristics into the optimization process, the proposed approach encourages the model to allocate greater learning capacity to decision-critical elements.

\subsection{Adjusted Loss Function Formulation}

We begin by formulating the adjusted loss function, deferring the specific construction of the BAA to the subsequent subsection.

The adjusted loss is defined as:
\begin{equation}
\label{LAdj}
L_{Adj,\delta}(Pred,Gt)=\sum_{i}(Adj(Pred_{i},Gt_{i})+\delta)L(Pred_{i},Gt_{i}),
\end{equation}
where $i$ indexes pixels, $L$ denotes a baseline pixel-wise loss such as WBCE\footnote{The same symbol $L$ is used to denote the loss on a single element $L(Pred_i,Gt_i)$ and on the entire prediction $L(Pred,Gt)$.}, $Pred$ and $Gt$ represent the model prediction and ground truth, respectively, and $\delta \geq 0$ is a constant added to stabilize training. The function $Adj$, referred to as an \textit{adjuster}, assigns dynamic weights in $[0,1]$ according to each element’s contribution to the final binary decision.

A straightforward but impractical choice of $Adj$ is the hard binary mask:
\begin{equation}
\label{Adjhat}
\hat{Adj}_{thr}(Pred_{i},Gt_{i})=\left\{\begin{matrix}
0 & \text{if } (Pred_{i}-thr)(Gt_{i}-thr)\geq 0,\\
1 & \text{otherwise},
\end{matrix}\right.
\end{equation}
where $thr$ denotes the binarization threshold. This formulation ensures that correctly classified elements incur zero loss, while misclassified ones incur positive loss, directly reflecting the binary decision. However, the resulting loss is discontinuous and unsuitable for gradient-based optimization, motivating the need for a continuous approximation.

\subsection{Design of the Binarization-Aware Adjuster}

Let $thr$ denote the binarization threshold. The adjuster $Adj$ is designed to satisfy the following conditions:

\begin{itemize}
\item \textbf{C1 (Continuity):} $Adj$ must be continuous to support gradient-based optimization.
\item \textbf{C2 (Non-negativity):} $Adj(Pred_{i},Gt_{i})\in[0,1]$, as it represents a weighting factor.
\item \textbf{C3 (Uncertainty Emphasis):} For correctly classified elements, the weight should decrease as the prediction goes away from $thr$.
\item \textbf{C4 (Confidence Saturation):} For confidently correct elements far from $thr$, the weight should rapidly decay, become zero beyond a predefined distance $thr\_dev$.
\item \textbf{C5 (Classification Sensitivity):} Misclassified elements should receive the maximal weight.
\end{itemize}

An adjuster satisfying these properties is referred to as a \textit{Binarization-Aware Adjuster (BAA)}.

\begin{Definition}
A \textbf{Binarization-Aware Adjuster (BAA)} with respect to $thr\in[0,1]$ and $thr\_dev>0$ is a function valued in $[0,1]$ satisfying conditions C1–C5.
\end{Definition}

To construct such an adjuster, we introduce the \textit{Distance Weight Function (DWF)}.

\begin{Definition}
A function $f_{thr\_dev}(x)$ is a \textbf{Distance Weight Function (DWF)} with threshold window $thr\_dev\in\mathbb{R}_{+}$ if it satisfies continuity, monotonic decrease, concavity, and boundary conditions:
$f_{thr\_dev}(0)=1$ and $f_{thr\_dev}(thr\_dev)=0$.
\end{Definition}

Using a DWF, an intermediate adjuster satisfying C1–C4 can be defined as:
\begin{equation}
\bar{Adj}(Pred_{i},Gt_{i})=\left\{\begin{matrix}
f_{thr\_dev}(|Pred_{i}-thr|), & |Pred_{i}-thr|\leq thr\_dev,\\
0, & |Pred_{i}-thr|>thr\_dev.
\end{matrix}\right.
\end{equation}

A concrete example of DWF is:
\begin{equation}
\label{DWF}
f_{thr\_dev}(x):=f_{b,thr\_dev}(x):=\frac{e^{b\times x}-e^{b\times thr\_dev}}{1-e^{b\times thr\_dev}},
\end{equation}
where $b\geq 0$ controls the decay rate.

To additionally satisfy C5, we define the masked distance:
\begin{Definition}
The ($[0,1]$) masked distance $MD_{thr}(x,y): [0,1]^{2}\longrightarrow [0,1]$ regarding the threshold $thr\in [0,1]$ is defined as: 
     \begin{equation}
     \label{MD}
    MD_{thr}(x,y)=(x-thr)\times y + (thr-x) \times (1-y)
\end{equation}
\end{Definition}

Hence, $MD_{thr}(Pred_{i},Gt_{i})$ measures both the correctness and confidence of the element $i$. For a wrongly predicted element $i$, namely whose prediction is on the different side of the $thr$ as its ground truth value, $MD_{thr}(Pred_{i},Gt_{i})\leq 0$, while for a correctly predicted element $i$, $MD_{thr}(Pred_{i},Gt_{i})$ is the $L^{1}$ distance of $Pred_{i}$ from the binarization threshold $thr$. 

The final BAA is then defined as:
\begin{equation}
\label{BAA}
BAA_{thr,thr\_dev}(Pred_{i},Gt_{i})=\hat{f}_{thr\_dev}(MD_{thr}(Pred_{i},Gt_{i})),
\end{equation}
where $\hat{f}_{thr\_dev}$ is an extension of a selected DWF $f_{thr\_dev}(x)$, defined as:
\begin{equation}
\label{EDWF}
\hat{f}_{thr\_dev}(x)=\left\{\begin{matrix}
1, & x<0,\\
f_{thr\_dev}(x), & 0\leq x\leq thr\_dev,\\
0, & x>thr\_dev.
\end{matrix}\right.
\end{equation}

\subsection{Theoretical Properties of the BAA}

To better understand the proposed adjuster’s behavior, we analyze its mathematical properties. Let $f_{thr\_dev}(x)=f_{b,thr\_dev}(x)$ be defined as in Equations \ref{DWF} and $\hat{f}_{thr\_dev}(x)=\hat{f}_{b,thr\_dev}(x)$ be its extension defined in Equation \ref{EDWF}. Then $\hat{f}_{b,thr\_dev}(x)$ serves as a smooth approximation to a hard binarization function. Specifically, it generalizes the discrete threshold-based binary weighting function $B_{thr\_dev}(x)$ on threshold $thr\_dev$, defined as:
\begin{equation}
    B_{thr\_dev}(x)=\left\{\begin{matrix}
 &1 &x<thr\_dev\\
&0\ &x\geq thr\_dev
\end{matrix}\right.
\end{equation}
The two functions coincide as $b\rightarrow+\infty$:
\begin{equation}
\lim_{b\rightarrow+\infty}\hat{f}_{b,thr\_dev}(x)=B_{thr\_dev}(x)
\end{equation}
Therefore, $\hat{f}_{b,thr\_dev}$ is a continuous approximation for threshold logic. Accordingly, the $BAA_{thr,thr\_dev}(Pred_{i},Gt_{i})=BAA_{thr,b,thr\_dev}(Pred_{i},Gt_{i})$, defined by Equation \ref{BAA}, converges to a discrete decision-based adjuster $\widetilde{Adj}_{thr,thr\_dev}(Pred_{i},Gt_{i})$, defined as:
\begin{equation}
\label{Bhat}
    \widetilde{Adj}_{thr,thr\_dev}(Pred_{i},Gt_{i})=\left\{\begin{matrix}
 &0 & (Pred_{i}-thr)(Gt_{i}-thr)\geq 0\ \&\ |Pred_{i}-thr|\geq thr\_dev\\
&1 &Otherwise
\end{matrix}\right.
\end{equation}
i.e.,
\begin{equation}
\lim_{b\rightarrow+\infty}BAA_{thr,b,thr\_dev}(Pred_{i},Gt_{i})=\widetilde{Adj}_{thr,thr\_dev}(Pred_{i},Gt_{i})
\end{equation}
This confirms that the BAA bridges the binarization logic with the continuous-loss gradient-based training. Furthermore, let also $thr\_dev\rightarrow 0^{+}$, then it is reduced to $\hat{Adj}_{thr}$ from Equation \ref{Adjhat}.

Moreover, for a DWF $f_{thr\_dev}(x)$, the following bounds holds:
\begin{equation}
    1-\frac{x}{thr\_dev}\leq f_{thr\_dev}(x)\leq 1
\end{equation} 
Therefore:
\begin{equation}
\lim_{thr\_dev\rightarrow+\infty}f_{thr\_dev}(x)=1
\end{equation}
which implies that:
\begin{equation}
\lim_{thr\_dev\rightarrow+\infty}BAA_{thr,thr\_dev}(x)=1
\end{equation}
and the adjusted loss $L_{Adj,\delta}$ defined in Equation \ref{LAdj} degenerates to the baseline loss $L$ up to a constant:
\begin{equation}
\lim_{thr\_dev \to \infty}L_{Adj,\delta}(Pred, Gt) = (1+\delta)L(Pred, Gt)
\end{equation}

In summary, the proposed adjusted loss function $L_{Adj,\delta}$ with $Adj=BAA_{thr,b,thr\_dev}$ defined above by the DWF in Equation \ref{DWF}, generalizes the baseline loss $L$. It degenerates to the baseline loss (up to a scale factor) when $thr\_dev\rightarrow+\infty$, converges to a binary weighted loss when $b\rightarrow+\infty$, and maintains continuity. These asymptotic behaviors confirm that the constructed BAA bridges the continuous loss and post-binarization classification, aligning training with downstream evaluation.

\subsection{Hyperparameter Selection and Training Protocol}
\label{TP}

\begin{figure}
    \centering
    \includegraphics[width=1\linewidth]{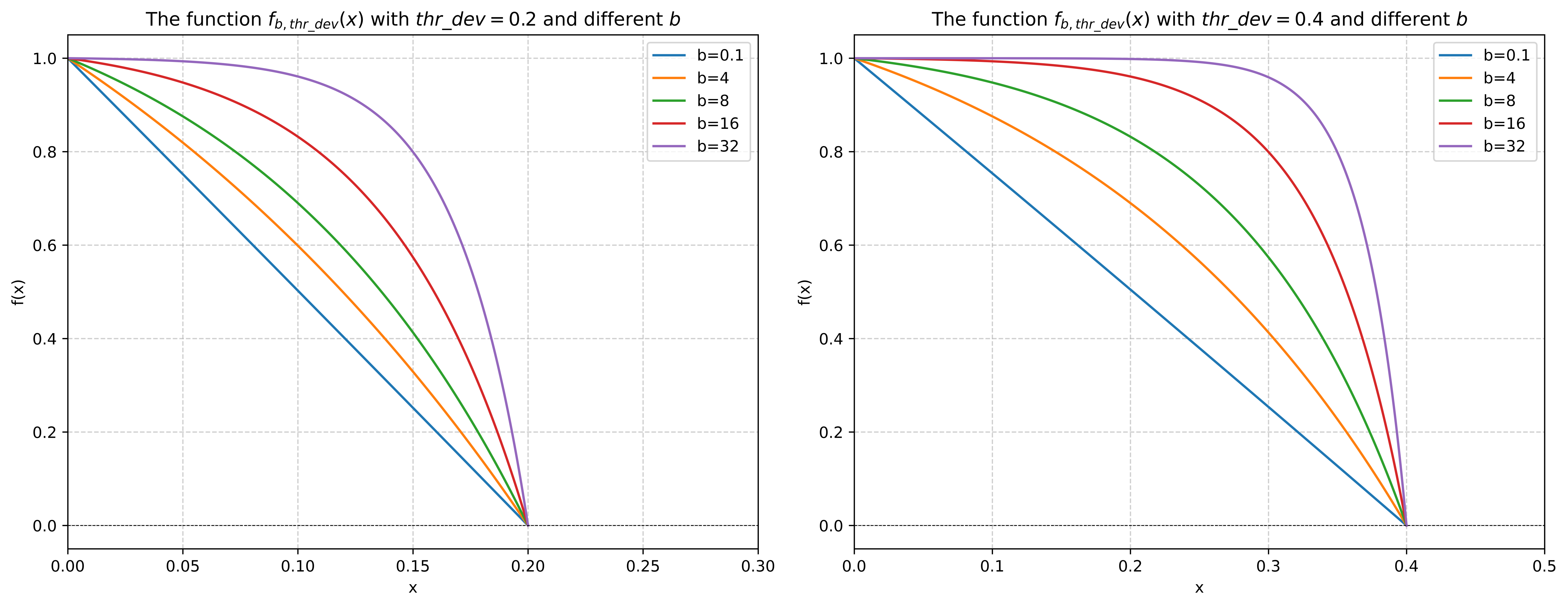}
    \caption{\ The different choices of $b$ in the DWF $f_{b,thr\_dev}(x)$ presented in Equation \ref{DWF}}
    \label{b}
\end{figure}

The proposed $BAA$ depends on three hyperparameters: $thr$: binarization threshold, $thr\_dev$: threshold window beyond which correct predictions are ignored in optimization, and $b$ (when employing $f_{b,thr\_dev}(x)$ from Equation \ref{DWF} as the DWF): DWF decay rate,

 The DWF decay rate $b$ occurs specifically in the DWF defined in Equation \ref{DWF}, controlling the concavity of $f_{b,thr\_dev}(x)$, determining how sharply the adjusted weight decays as predictions diverge from the decision boundary $thr$. Its effect is illustrated in Figure \ref{b}, and we set $b=16$ as the default.

$thr\_dev$ defines the margin around $thr$ within which a correct prediction is considered uncertain. A correct prediction with $|Pred_{i}-thr|>thr\_dev$ is considered as confidently correct and receives zero weight. We empirically choose $thr\_dev=0.2$ for the ED task.

$thr$ is the most critical hyperparameter in the BAA. For ED models, experiments (Table \ref{Thr}) show that the optimal value typically lies between 0.6 and 0.8, so we use $thr=0.7$ as the default.

However, given its importance, it is worth providing a calibration method for it. Specifically, we suggest a self-adaptive $thr$ chosen by pre-training: pre-train the model on a sub-dataset without applying BAA and evaluate the pre-trained model on a validation set to select a suitable $thr$ for BAA.

Therefore, the final training protocol of a model $M$ under a selected baseline loss function $L$ on a dataset $D$, with the presented BAA method, can be formulated as:

\begin{itemize}
    \item Step 1: Split $D$ into pre-train set $P$, validation set $V$, and test set $T$.
    \item Step 2: Train $M$ on $P$ using baseline loss $L$, and evaluate on $V$ to estimate $thr$.
    \item Step 3: Retrain $M$ via $L_{BAA,\delta}$, on $P\cup V$, and evaluate finally on $T$.
\end{itemize}

In experiments, $\delta$ is simply set to be 1.

\section{Experiment: Application to Edge Detection}
\label{Experiment}

To validate the effectiveness of the proposed BAA framework, we apply it to the ED task. ED is selected as it is a fundamental problem in computer vision, serving as a core component for many higher-level vision tasks and providing a clear setting for evaluating binary decision learning.

\subsection{Experimental Setting}
\label{Settings}

We use weighted binary cross-entropy (WBCE) as the baseline loss function $L$, due to its widespread adoption in ED tasks, and adopt the adjuster defined in Equations~\ref{DWF}, \ref{MD}, \ref{BAA}, and \ref{EDWF} as the BAA.

\subsubsection{Baseline and Benchmark}

The proposed method is evaluated on a representative set of ED models, including HED (2015)~\cite{XT2015Holistically}, BDCN (2022)~\cite{HZ2022BDCN}, Dexi (2023)~\cite{SS2023Dense}, and EES3-Double (2025)~\cite{S2025Boost}, where EES3-Double denotes the EES framework integrating the above three models with a double-size selector.

Experiments are conducted on four benchmark datasets: BRIND \cite{PH2021RINDNet}, BSDS500 \cite{MF2001A}, UDED \cite{SL2023Tiny}, and BIPED2 \cite{SR2020Dense}.

All models are retrained from scratch on each dataset to ensure fairness. For consistency and simplicity, only the final output of each model is used for supervision, bypassing any need for hyperparameter tuning in intermediate layers. Each model is trained under the following three loss configurations for comparison:
\begin{itemize}
    \item Standard WBCE (baseline).
    \item WBCE with a fixed-threshold BAA using $thr=0.7$ (denoted as \textit{BAA-0.7})
    \item WBCE with a self-adaptive adjuster where $thr$ is automatically selected following the method in Section \ref{TP} (denoted as \textit{BAA-SA}).
\end{itemize}

Unless otherwise specified, the hyperparameters for the adjusted loss defined by the BAA are fixed at $thr\_dev=0.2$, $b=16$, and $\delta=1$.

Evaluation protocols follow the standard ED metrics (ODS and OIS) introduced in \cite{MF2004Learning}, under the strictest 1-pixel error tolerance without post-processing techniques such as non-maximal suppression (NMS), suggested in \cite{S2025Enhancing}. The evaluations under the traditional relaxed error tolerance (4.3 to 11.1 pixels) with NMS are provided in the appendix.

\subsubsection{Dataset Partitions and Augmentation}

Dataset preprocessing and postprocessing procedures follow those introduced in \cite{S2025Boost}. The details are summarized as follows.

The dataset partitions follow: BRIND and BSDS500: 400 training images and 100 testing images, UDED: 20 training images and 7 testing images (excluding two images with resolution below $320\times 320$), and BIPED2: 200 training images and 50 testing images.

For training under the self-adaptive adjuster setting in section \ref{TP}, each training set is further divided into pre-training and validation subsets, following: BRIND and BSDS500: 300 pre-training and 100 validation images, UDED: 15 pre-training and 5 validation images, and BIPED2: 150 pre-training and 50 validation images.

Data augmentations are applied, including: Halving the images until both height and width are strictly below 640 pixels, rotations at four angles, $0^{\circ}$, $90^{\circ}$, $180^{\circ}$, and $270^{\circ}$, with horizontal flips, as well as addition of noiseless data.

\subsubsection{Training Note}

During training, input images are randomly cropped to $320\times320$ and refreshed every 5 epochs. The batch size is set to 8. All models are trained on an A100 GPU using Adam optimizer, with a learning rate of 0.0001 and a weight decay of $10^{-8}$.

Training runs for 200 epochs on UDED and 50 epochs on the remaining datasets.

\subsubsection{Inference Method}

In the inference phase, each image is split into overlapping $320\times 320$ patches with a stride of 304 pixels (i.e., 16 pixels overlap). The model predicts edge maps on each patch independently, and the outputs are aggregated to form the final edge map.

Further implementation details can be found in the code or in \cite{S2025Boost}.

\subsection{Experimental Results}

Tables \ref{BRIND-BSDS} and \ref{UDED-BIPED2} report the ED performance of all methods on BRIND, BSDS500, UDED, and BIPED2 datasets under the strict 1-pixel error tolerance setting, without applying NMS. The evaluation includes ODS and OIS scores for each configuration.

\begin{table}[htbp]
\renewcommand\arraystretch{1.15}
\centering
\caption{\textbf{Results on BRIND and BSDS500 under 1-pixel error tolerance without NMS}: For BRIND, all annotations per image are merged into a single binary ground truth map, where a pixel is considered an edge one if it is marked as such by any annotator. For BSDS500, only label 1 is used as the ground truth. Notations follow those introduced in the main-text. The best scores for each model are highlighted in \textbf{bold}. Rows labeled \textit{A-No-BAA}, \textit{A-BAA-0.7}, and \textit{A-BAA-SA} represent the average performance across the four models corresponding to each setting, respectively, while \textit{Improvement} rows indicate the relative improvements over the baseline.}
\label{BRIND-BSDS}
\begin{tabular}{|p{40mm}<{\centering}|p{32.35mm}<{\centering}|p{32.35mm}<{\centering}|}
\hline
 & BRIND & BSDS500
\end{tabular}
\begin{tabular}{|p{40mm}<{\centering}|p{14mm}<{\centering}|p{14mm}<{\centering}|p{14mm}<{\centering}|p{14mm}<{\centering}|}
\hline
    & ODS   & OIS & ODS   & OIS\\
\hline
HED (2015) & 0.664 & 0.673 & 0.444 & 0.455\\
\hline
HED-BAA-0.7 & \textbf{0.668} & \textbf{0.676} & 0.441 & 0.453\\
\hline
HED-BAA-SA & 0.664 & 0.672 & \textbf{0.455} & \textbf{0.465}\\
\hline
BDCN (2022) & 0.655 & 0.663 & 0.432 & 0.446 \\
\hline
BDCN-BAA-0.7  & 0.662 & 0.671 & 0.445 & 0.452\\
\hline
BDCN-BAA-SA & \textbf{0.665}& \textbf{0.674}& \textbf{0.454} & \textbf{0.468} \\
\hline
Dexi (2023) & 0.675 & 0.683& 0.465 & 0.473\\
\hline
Dexi-BAA-0.7 & 0.674& 0.683& \textbf{0.475} & \textbf{0.485}\\
\hline
Dexi-BAA-SA & \textbf{0.679}& \textbf{0.687}& 0.470 & 0.480\\
\hline
EES3-Double (2025) & 0.683 & 0.690& 0.488 & 0.504\\
\hline
EES3-Double-BAA-0.7 & \textbf{0.690}& \textbf{0.697}& 0.493 & \textbf{0.509}\\
\hline
EES3-Double-BAA-SA & \textbf{0.690}& \textbf{0.697}& \textbf{0.497} & 0.507\\
\hline
A-No-BAA & 0.669 & 0.677 & 0.457 & 0.470 \\
\hline
A-BAA-0.7 & 0.674 & 0.682 & 0.464 & 0.475 \\
\hline
A-BAA-0.7-Improvement & +0.75$\%$ & +0.74$\%$ & +1.53$\%$ & +1.06$\%$ \\
\hline
A-BAA-SA & \textbf{0.675} & \textbf{0.683} & \textbf{0.469} & \textbf{0.480} \\
\hline
A-BAA-SA-Improvement & \textbf{+0.90$\%$} & \textbf{+0.89$\%$} & \textbf{+2.63$\%$} &\textbf{+2.13$\%$} \\
\hline
\end{tabular}
\end{table}

\begin{table}[htbp]
\renewcommand\arraystretch{1.15}
\centering
\caption{\textbf{Results on UDED and BIPED2 under 1-pixel error tolerance without NMS}: Notations follow the ones in Table\ref{BRIND-BSDS}.}
\label{UDED-BIPED2}
\begin{tabular}{|p{40mm}<{\centering}|p{32.35mm}<{\centering}|p{32.35mm}<{\centering}|}
\hline
 & UDED & BIPED2
\end{tabular}
\begin{tabular}{|p{40mm}<{\centering}|p{14mm}<{\centering}|p{14mm}<{\centering}|p{14mm}<{\centering}|p{14mm}<{\centering}|}
\hline
    & ODS   & OIS & ODS   & OIS  \\
\hline
HED (2015) & 0.709 & 0.742& 0.630 & 0.633 \\
\hline
HED-BAA-0.7  & \textbf{0.727} & \textbf{0.755} & 0.629 & 0.634 \\
\hline
HED-BAA-SA & 0.712 & 0.748 & \textbf{0.633} & \textbf{0.635}\\
\hline
BDCN (2022) & 0.688 & 0.723& 0.636 & 0.639 \\
\hline
BDCN-BAA-0.7  & \textbf{0.713} & 0.741 & 0.637 & 0.639 \\
\hline
BDCN-BAA-SA & 0.712 & \textbf{0.749} & \textbf{0.640} & \textbf{0.643}\\
\hline
Dexi (2023) & 0.700 & 0.738 & \textbf{0.643} & \textbf{0.648}\\
\hline
Dexi-BAA-0.7  & 0.721 & 0.752 & 0.640 & 0.646 \\
\hline
Dexi-BAA-SA & \textbf{0.731} & \textbf{0.765} & 0.641 & 0.646\\
\hline
EES3-Double (2025)  & 0.751 & 0.780& 0.670 & 0.673 \\
\hline
EES3-Double-BAA-0.7  & \textbf{0.762} & \textbf{0.794} & 0.671 & 0.675 \\
\hline
EES3-Double-BAA-SA & 0.758 & 0.791 & \textbf{0.675} & \textbf{0.680}\\
\hline
A-No-BAA & 0.712 & 0.746 & 0.645 & 0.648 \\
\hline
A-BAA-0.7 & \textbf{0.731} & 0.761 & 0.644 & 0.649 \\
\hline
A-BAA-0.7-Improvement &\textbf{+2.67$\%$} & +2.01$\%$ &-0.16$\%$ &+0.15$\%$ \\
\hline
A-BAA-SA & 0.729 & \textbf{0.763} & \textbf{0.647} & \textbf{0.651} \\
\hline
A-BAA-SA-Improvement & +2.39$\%$ &\textbf{+2.28$\%$} & \textbf{+0.31$\%$} &\textbf{+0.46$\%$} \\
\hline
\end{tabular}
\end{table}

The results indicate that incorporating the proposed BAA method, either with a fixed threshold (\textit{BAA-0.7}) or with the self-adaptive setting (\textit{BAA-SA}), generally leads to improved performance across most evaluated models and datasets. Both variants of the BAA loss outperform the baseline WBCE loss in the majority of evaluated settings. The self-adaptive variant (\textit{BAA-SA}) achieves slightly higher scores than the fixed-threshold version in several cases.

On BRIND, the best ODS gains are observed in BDCN (about +1.5$\%$) using the self-adaptive adjuster (\textit{BAA-SA}). On BSDS500, EES3-Double and BDCN benefit most from the adjuster, with ODS improvements up to +1.8$\%$ and +2.6$\%$, respectively. UDED shows the most significant performance gains. The average ODS improves by about +2.7$\%$ (\textit{BAA-0.7}) and +2.4$\%$ (\textit{BAA-SA}) compared to the baseline. On BIPED2, the improvements are modest but consistent with \textit{BAA-SA} outperforming both baseline and fixed-threshold setups.

Overall, these results support the effectiveness of the proposed BAA method in the context of ED. The performance gains are consistent across different backbone architectures and datasets, while requiring no modification to model architectures or inference procedures.

\subsection{Ablation Study}

\begin{figure}
    \centering
    \includegraphics[width=\linewidth]{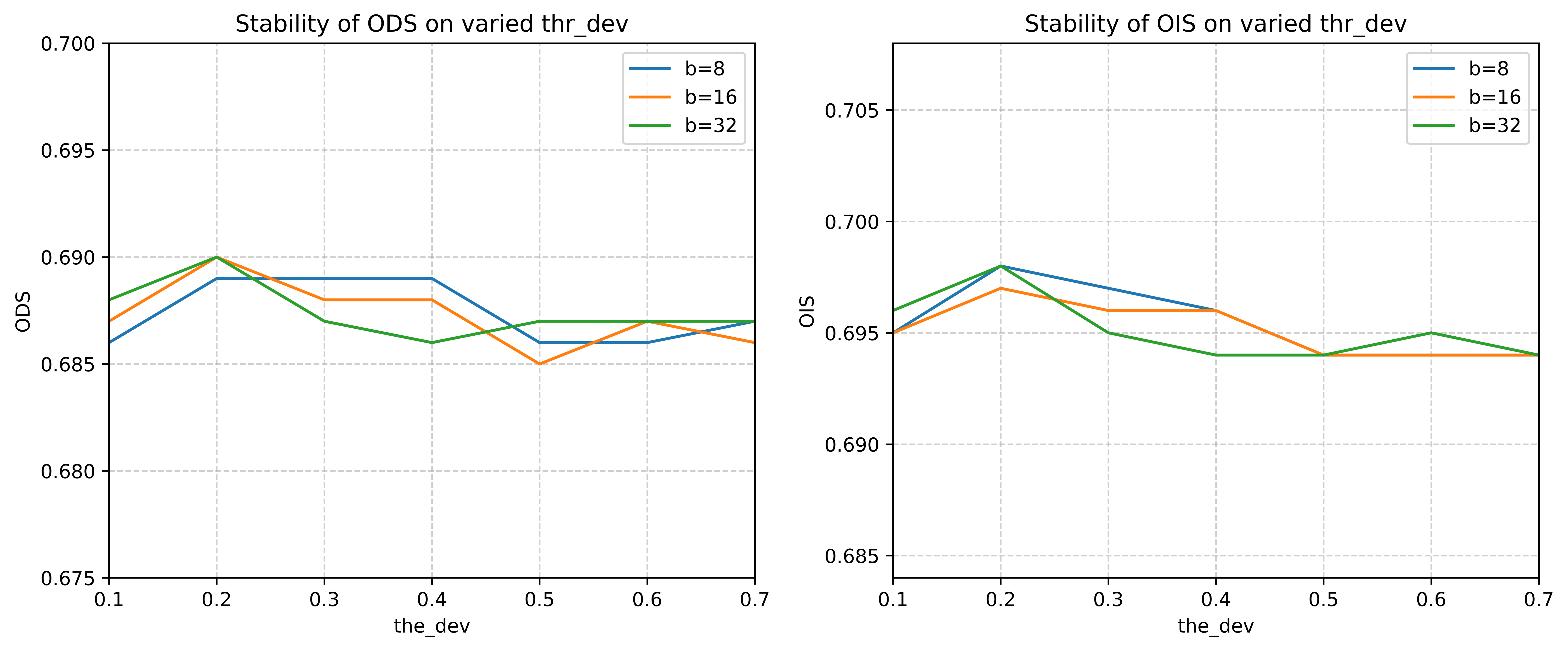}
    \caption{\ \textbf{ODS and OIS across $thr\_dev$}: ODS and OIS values under different $thr\_dev$ settings (0.1 to 0.7) and decay rates $b$ (8, 16, 32). Experiments are conducted on BRIND using the EES3-Double model with the adjusted WBCE loss. The threshold $thr$ is fixed at 0.7 and $\delta$ at 1.}
    \label{Stability}
\end{figure}

Figure \ref{Stability} illustrates how ODS and OIS scores vary with different values of $thr\_dev$ (from 0.1 to 0.7) and decay rates $b$ (8, 16, 32) in the DWF (Equation \ref{DWF}). All experiments are conducted on the BRIND dataset using the EES3-Double model trained with the BAA-WBCE loss. The threshold $thr$ is fixed at 0.7, and $\delta$ at 1.

This ablation study shows that across all tested configurations, $thr\_dev = 0.2$ consistently yields the best performance. Moreover, the proposed BAA method is robust to a wide range of hyperparameter settings, suggesting that it is not overly sensitive to hyperparameter choices in the ED setting and that a unified configuration can be adopted in practice.

\section{Discussion}

Although the BAA is formulated for binary decision tasks, its underlying principle is not restricted to binarization itself. Conceptually, the framework can be extended to multi-valued discrete inference by applying the adjuster around individual decision boundaries. In such cases, when a correct prediction goes away from any discrete boundary, a decreased optimization weight can be assigned accordingly. From a theoretical perspective, this extension can even be realized by the binary formulation established in Section~\ref{Methodology}, as each local decision boundary can be treated as a binary decision problem.

In this work, experimental validation is conducted using ED as a representative task. This choice is motivated by the clear binary inference rule in ED and its suitability for analyzing optimization-inference misalignment. The primary goal of this study is to establish a theoretically grounded adjustment mechanism rather than to provide an exhaustive empirical evaluation across tasks. While ED also underlies several related vision problems, we deliberately limit our experiments to this setting to maintain a focused and controlled analysis. Further empirical investigations on other tasks are therefore left as future work.

One practical consideration of the BAA formulation is that additional model training for the binarization threshold estimation introduces additional training cost, despite not affecting inference efficiency. However, this overhead is not inherent to the BAA framework itself, but rather stems from the fact that the optimal decision threshold is not accessible in practical tasks. From this perspective, the two-stage estimation process should also be regarded as a general-purpose strategy for approximating decision thresholds in discrete inference problems, rather than as a limitation specific to the BAA framework.

Another limitation is that the estimated threshold may not coincide exactly with the true optimal inference threshold. For instance, in ED, the optimal threshold under the ODS metric is determined using test data, which is inherently inaccessible during training. As a result, the self-adaptive threshold should be interpreted as an effective surrogate rather than an exact optimum. Nevertheless, such a surrogate provides a principled and practically feasible approximation for guiding binarization-aware optimization.

Overall, despite the need for threshold calibration and the focus on a single task, the proposed framework establishes a principled foundation for improving the alignment between continuous optimization and discrete inference in decision-based learning problems.

\section{Conclusion}

This paper addresses a fundamental issue in discrete decision learning: the misalignment between continuous-valued optimization during training and discrete predictions at inference, which arises from the discontinuity of discretization operations that are excluded from standard learning pipelines. To address this issue, we introduce a theoretically grounded binarization-aware adjustment (BAA) mechanism, guided by the Distance Weight Function (DWF). BAA incorporates binarization characteristics into continuous optimization by modulating loss contributions according to prediction correctness and proximity to the decision threshold, emphasizing decision-critical elements and downweighting those with limited influence. The proposed framework is applied to the edge detection (ED) task, where binary inference is misaligned with the standard model training. Experimental results on representative models and datasets show that adopting BAA into optimization consistently improves performance. Overall, the BAA provides a principled mechanism for bridging continuous optimization and discrete evaluation, offering a theoretically sound direction for improving decision-aligned learning in discrete inference tasks.

\bibliographystyle{unsrt}
\bibliography{EDBitex}

\section*{Data availability}

 Codes could be found at https://github.com/Hao-B-Shu/ES-EES-SWBCE-EBT-BAA.





\section*{Declaration of Interests}

The author(s) declare no conflict of interest.

\section{Supplementary Material}

\subsection{Evaluations on Relaxed Benchmarks in ED tasks}

Tables \ref{BRIND-BSDS-Old} and \ref{UDED-BIPED2-Old} present the evaluation results under the traditional relaxed criterion, with the pixel error tolerance ranging from 4.3 to 11.1 pixels, depending on the resolution of the evaluated image, and using NMS.

\begin{table}[htbp]
\renewcommand\arraystretch{1.15}
\centering
\caption{\textbf{Results on BRIND and BSDS500 with 4.3-pixel error tolerance and NMS}: Notations follow those used in the main-text.}
\label{BRIND-BSDS-Old}
\begin{tabular}{|p{40mm}<{\centering}|p{32.35mm}<{\centering}|p{32.35mm}<{\centering}|}
\hline
 & BRIND & BSDS500
\end{tabular}
\begin{tabular}{|p{40mm}<{\centering}|p{14mm}<{\centering}|p{14mm}<{\centering}|p{14mm}<{\centering}|p{14mm}<{\centering}|}
\hline
    & ODS   & OIS& ODS   & OIS\\
\hline
HED (2015) & 0.781 & 0.795& 0.613 & 0.630\\
\hline
HED-BAA-0.7& \textbf{0.786} & \textbf{0.798}& 0.618 & 0.639\\
\hline
HED-BAA-SA & 0.781 & 0.793& \textbf{0.629} & \textbf{0.642}\\
\hline
BDCN (2022) & 0.770 & 0.783& 0.609 & 0.631 \\
\hline
BDCN-BAA-0.7  & 0.778& \textbf{0.791}& 0.618 & 0.634 \\
\hline
BDCN-BAA-SA & \textbf{0.779} & \textbf{0.791}& \textbf{0.622} & \textbf{0.645} \\
\hline
Dexi (2023) & 0.788 & 0.800& 0.642 & 0.663 \\
\hline
Dexi-BAA-0.7 & \textbf{0.791}& \textbf{0.804} & \textbf{0.648} & \textbf{0.673 }\\
\hline
Dexi-BAA-SA & \textbf{0.791}& 0.803 & 0.639 & 0.651\\
\hline
EES3-Double (2025) & 0.792 & 0.804& 0.651 & \textbf{0.680} \\
\hline
EES3-Double-BAA-0.7 & \textbf{0.797}& \textbf{0.809}& 0.649 & 0.670 \\
\hline
EES3-Double-BAA-SA & \textbf{0.797}& 0.808& \textbf{0.654} & 0.663 \\
\hline
A-No-BAA & 0.683 & 0.796 & 0.629 & 0.651 \\
\hline
A-BAA-0.7 & \textbf{0.688} & \textbf{0.801} & 0.633 & \textbf{0.654} \\
\hline
A-BAA-0.7-Improvement & \textbf{+0.73$\%$} & \textbf{+0.63$\%$} & +0.64$\%$ & \textbf{+0.46$\%$} \\
\hline
A-BAA-SA & 0.687 & 0.799 & \textbf{0.636} & 0.650  \\
\hline
A-BAA-SA-Improvement & +0.59$\%$ & +0.38$\%$ &\textbf{+1.11$\%$} &-0.15$\%$ \\
\hline
\end{tabular}
\end{table}

\begin{table}[htbp]
\renewcommand\arraystretch{1.15}
\centering
\caption{\textbf{Results on UDED and BIPED2 with 0.0075 error tolerance and NMS}: Notations follow those used in the main-text. A 0.0075 error tolerance corresponds to approximately 11.1-pixel tolerance for BIPED2.}
\label{UDED-BIPED2-Old}
\begin{tabular}{|p{40mm}<{\centering}|p{32.35mm}<{\centering}|p{32.35mm}<{\centering}|}
\hline
 & UDED & BIPED2
\end{tabular}
\begin{tabular}{|p{40mm}<{\centering}|p{14mm}<{\centering}|p{14mm}<{\centering}|p{14mm}<{\centering}|p{14mm}<{\centering}|}
\hline
    & ODS   & OIS & ODS   & OIS\\
\hline
HED (2015) & 0.806 & 0.846& 0.883 & 0.890 \\
\hline
HED-BAA-0.7& \textbf{0.818} & \textbf{0.849}& \textbf{0.884} & \textbf{0.892} \\
\hline
HED-BAA-SA & 0.807 & 0.843& 0.883 & 0.890 \\
\hline
BDCN (2022) & 0.786 & 0.822& 0.883 & 0.890 \\
\hline
BDCN-BAA-0.7 & 0.798 & 0.828& 0.881 & 0.889 \\
\hline
BDCN-BAA-SA & \textbf{0.801} & \textbf{0.832}& \textbf{0.886} & \textbf{0.893} \\
\hline
Dexi (2023) & 0.788 & 0.824& \textbf{0.887} & \textbf{0.895} \\
\hline
Dexi-BAA-0.7  & 0.801 & 0.831& 0.884 & 0.892 \\
\hline
Dexi-BAA-SA & \textbf{0.814} & \textbf{0.858} & 0.886 & 0.894\\
\hline
EES3-Double (2025)  & 0.831 & 0.858& 0.888 & 0.896 \\
\hline
EES3-Double-BAA-0.7  & \textbf{0.834} & \textbf{0.863} & 0.889 & 0.897 \\
\hline
EES3-Double-BAA-SA & 0.832 & 0.859 & \textbf{0.890} & \textbf{0.898}\\
\hline
A-No-BAA & 0.803 & 0.838 & 0.885 & 0.893 \\
\hline
A-BAA-0.7 & 0.813 & 0.843 & 0.885 & \textbf{0.894} \\
\hline
A-BAA-0.7-Improvement & +1.25$\%$ & +0.60$\%$ &+0.00$\%$ &\textbf{+0.11$\%$} \\
\hline
A-BAA-SA & \textbf{0.814} & \textbf{0.848} & \textbf{0.886} & \textbf{0.894} \\
\hline
A-BAA-SA-Improvement &\textbf{+1.37$\%$} &\textbf{+1.19$\%$} &\textbf{+0.11$\%$} & \textbf{+0.11$\%$} \\
\hline
\end{tabular}
\end{table}

\subsection{Full Data Used in the Ablation Study}

Table \ref{Hyper-test} provides the detailed numerical data used to generate Figure \ref{Stability} in the ablation study.

\begin{table}[htbp]
\renewcommand\arraystretch{1.1}
\centering
\caption{\textbf{Hyperparameter Evaluation for $thr\_dev$ and $b$}: Results are reported on the BRIND dataset under 1-pixel error tolerance without NMS, corresponding to Figure \ref{Stability}. The table lists ODS and OIS scores for various combinations of $thr\_dev$ and $b$, while keeping $thr = 0.7$ and $\delta = 1$. Notations follow those in the main-text.}
\label{Hyper-test}
\begin{tabular}{|p{88.75mm}<{\centering}|}
\hline
  BRIND with 1-pixel error tolerance without NMS   
\end{tabular}
\begin{tabular}{|p{50mm}<{\centering}|p{15mm}<{\centering}|p{15mm}<{\centering}|}
\hline
  Hyperparameter: $thr$-$thr\_dev$-$\delta$-$b$  & ODS   & OIS\\
\hline
EES3-Double-Adj: 0.7-0.1-1-8 & 0.686 & 0.695 \\
\hline
EES3-Double-Adj: 0.7-0.2-1-8 & 0.689& 0.698 \\
\hline
EES3-Double-Adj: 0.7-0.3-1-8 & 0.689 & 0.697 \\
\hline
EES3-Double-Adj: 0.7-0.4-1-8 & 0.689 & 0.696 \\
\hline
EES3-Double-Adj: 0.7-0.5-1-8 & 0.686 & 0.694 \\
\hline
EES3-Double-Adj: 0.7-0.6-1-8 & 0.686 & 0.694 \\
\hline
EES3-Double-Adj: 0.7-0.7-1-8 & 0.687 & 0.694 \\
\hline
EES3-Double-Adj: 0.7-0.1-1-16 & 0.687 & 0.695 \\
\hline
EES3-Double-Adj: 0.7-0.2-1-16 & 0.690 & 0.697 \\
\hline
EES3-Double-Adj: 0.7-0.3-1-16 & 0.688 & 0.696 \\
\hline
EES3-Double-Adj: 0.7-0.4-1-16 & 0.688 & 0.696 \\
\hline
EES3-Double-Adj: 0.7-0.5-1-16 & 0.685 & 0.694 \\
\hline
EES3-Double-Adj: 0.7-0.6-1-16 & 0.687 & 0.694 \\
\hline
EES3-Double-Adj: 0.7-0.7-1-16 & 0.686 & 0.694 \\
\hline
EES3-Double-Adj: 0.7-0.1-1-32 & 0.688 & 0.696 \\
\hline
EES3-Double-Adj: 0.7-0.2-1-32 & 0.690 & 0.698 \\
\hline
EES3-Double-Adj: 0.7-0.3-1-32 & 0.687 & 0.695 \\
\hline
EES3-Double-Adj: 0.7-0.4-1-32 & 0.686 & 0.694 \\
\hline
EES3-Double-Adj: 0.7-0.5-1-32 & 0.687 & 0.694 \\
\hline
EES3-Double-Adj: 0.7-0.6-1-32 & 0.687 & 0.695 \\
\hline
EES3-Double-Adj: 0.7-0.7-1-32 & 0.687 & 0.694 \\
\hline
\end{tabular}
\end{table}

\end{document}